# Towards Industrialized Conception and Production of Serious Games

I. Marfisi-Schottman, A. Sghaier, S. George, F. Tarpin-Bernard and P. Prévôt

*Abstract*—Serious Games (SGs) have experienced a tremendous outburst these last years. Video game companies have been producing fun, user-friendly SGs, but their educational value has yet to be proven. Meanwhile, cognition research scientist have been developing SGs in such a way as to guarantee an educational gain, but the fun and attractive characteristics featured often would not meet the public's expectations. The ideal SG must combine these two aspects while still being economically viable. In this article, we propose a production chain model to efficiently conceive and produce SGs that are certified for their educational gain and fun qualities. Each step of this chain will be described along with the human actors, the tools and the documents that intervene.

## I. INTRODUCTION

THESE past decades, computer games and web applications have become more and more familiar to people, whatever their age, gender or social environment. By taking advantage of this trend, the educational system has evolved to create an innovative generation of learning technology: Serious Games (SGs) [1]. An American study [2] revealed that, on average, students graduating with a bachelor's degree have spent only 5,000 h reading for over 10,000 h playing video games and 20,000 h watching TV!

Games have always been used to develop intelligence and build knowledge. Why would they not systematically be used for learning even with adults [3]? Games have the power to immerge learners into a world where they have to invest themselves intellectually and mentally to progress, face challenges or accomplish quests. The objective of a SG is to teach; the game dimension only serves to enhance the effectiveness by catalyzing the learner's attention. SGs add a new dimension to teaching by combining the two fundamental pillars of education: action and emotion. In traditional learning, only the parts of the brain dedicated to problem resolution, vision and speech are used, whereas SGs also stimulate the emotional, and sometimes, physical zones as well. SGs have two other undeniable advantages: the power to adapt to the learner and unlimited patience. The scenario, the level of difficulty, the speed of progression, and the nature of the internal elements of the game… everything can be adapted to the knowledge and preferences of the learners, or to their handicaps.

Riding the wave of user-friendly, intuitive home and portable consoles such as the Nintendo Wii [4] and DS [5], a wide range of commercial SGs have been developed. These SGs, developed as mini reflex games [6] or logical enigmas [7] are supposed to train reflexes and develop memory and logic faculties. Even though their fun qualities are widely recognized, their educational value has yet to be proven. Meanwhile, cognition research scientists have been developing their SGs in such a way as to guarantee an educational gain. However, the fun, attractive characteristics they feature are often hampered by old-fashioned graphics and monotonous interactions. How can we bring together the best of these two worlds to design SGs that are certified for their educational gain and fun qualities? In addition, to have commercially viable SGs, their conception and production needs to be rapid and efficient. This is all the more important knowing that the market span of SGs, due to their specific educational goals, are often very limited, thereby reducing the potential profits as compared to video games.

## II. MAIN ISSUE

SG production is a long process involving multiple actors. No specific protocol has yet been developed resulting in the use of methods designed for video games (on the one hand) or educational systems (on the other hand), that are then patched up in an effort to adapt them to the needs of SGs. The production starts by a conception phase during which domain experts, pedagogical experts and cognitive experts build a mock-up model of the future SG. The mock-up model is then transferred to the production team (programmers, graphic designers, actors, sound managers…) that build the first version of the SG called a prototype. This step is very time-consuming, all the more so as it is impossible to invest in mass production the way big video companies do. The prototype is then tested on a representative test group. This phase can go on for months or even years. The collected feedback triggers an important amount of recoding work to correct the errors and, sometimes, even the conception of the scenario needs to be rethought. In this last case, the SG will have to be resubmitted to the conception team and the production team.

For all these reasons, and based on our conception and production experience with a dozen SGs [8], we can say that the whole process lasts about three years (to obtain approximately 10h of learning) and the total cost comes out to approximately 15,000€ per hour of SG. These figures were also confirmed by several video companies. To minimize the constraints related to money and time, we must come up with a rapid and efficient production chain for SG conception and production.

The SG scientific community is probably now in a position to meet this challenge. Over these past fifteen years, researchers have acquired enough experience in the production of SGs to share their techniques and pursue new ambitions [9] [10]. SGs are becoming more and more popular due to the fact that they apply to new domains and therefore captivate the interest of a number of scientific and professional communities. It is the ideal moment to propose efficient and effective production methods for future SGs. It would also be wise to collaborate with video game companies and, in general, with all industrial companies that are used to dealing with time and money constraints.

To efficiently produce SGs that are certified for their educational value and fun qualities, we propose, in this article, a production chain model. In the next section, each step of this chain will be described along with the human actors, the tools and the documents that intervene.

### III. PRODUCTION CHAIN

#### A. Global view

To show all the different factors involved, we decided to apply the 5M classification, often used in industrial engineering: Method, Milieu, Manpower, Machine and Material [11]:

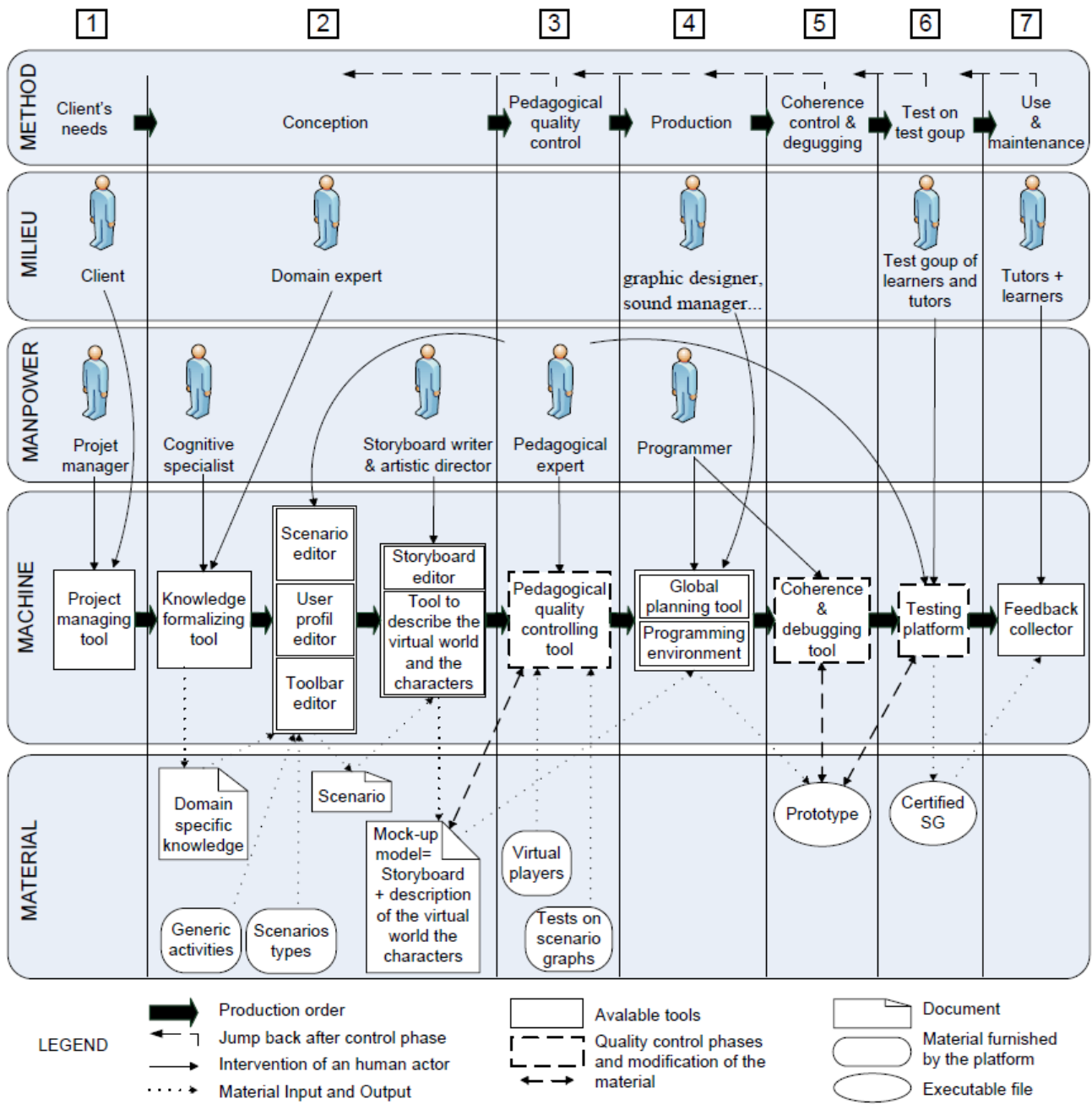

Fig. 1 Global vision of the SG production chain.

- Method: overall organization of the different production steps also including the flow of material input and output and the intervention of human actors.
- Milieu: all the elements of the external environment that intervene on the production of the SG such as domain experts (professors, doctors, engineers…), freelance subcontractors (graphic designers, actors, sound managers…) or the learners and the tutors (tests and feedback).
- Manpower: the team of human actors that is employed to work on the production chain. To facilitate comprehension, these actors will be described by their function (pedagogical expert, programmer…) although these various roles may in fact be detained by a single person.
- Machine: set of tools that will assist the human actors to produce the SG.
- Material: documents, mock-up models, executable files, data bases and all other artifacts that will be used as material to manufacture the final SG.

Fig. 1 illustrates the different steps of the SG production chain by using the previous classification. The next section describes the different steps in chronological order.

*B. General description of the production chain*

The production process starts with the client's request for a SG that meets his specific needs (Fig. 1, phase 1). The project manager is in charge of making sure the different steps of production are followed and done according to the cost and time constraints set by the client.

The next phase is the conception of a mock-up model of the SG and its pedagogical quality control (Fig. 1, phase 2 and 3). As these two phases are the center of our research, will be described them in detail in section IV.

The mock-up model is then passed on to the production team (Fig. 1, phase 4) composed of programmers and freelance specialists such as graphic designers, actors, film directors, sound managers; etc. All the members of this team collaborate together thanks to a global planning tool that helps them synchronize their work, communicate and share documents. After this phase, the SG prototype is then validated by a coherence control and debugging phase (Fig. 1, phase 5).

The prototype is now ready to be trialed on a representative test group (Fig. 1, phase 6).. This phase is certainly very time and money consuming but it is still essential. We can nonetheless make it much more efficient by using a testing platform that can automatically collect the tracking data of the tests and analyze them with Data mining techniques. After having spotted any existing problems, the pedagogical expert must resolve them by notifying the responsible members of the production team.

Once the SG has gone through all the preceding steps, it can be certified for its pedagogical value and fun qualities. The person in charge of the SG sessions (tutor or student) can then create accounts, plan and set up group playing sessions and perhaps even personalize certain parameters if the SG design allows it (Fig. 1, phase 7). If problems are encountered by the tutors or learners, they must be able to contact the production or conception team. Most SGs are designed to be online applications so they will support maintenance setups. It is also possible to equip the SG with a hidden tracking system to extract quality indicators, allowing the SG to be improved with time.

IV. CONCEPTION AND PEDAGOGICAL QUALITY CONTROL

A. *Extracting the domain-specific knowledge*

The first step of the conception phase consists in extracting the domain-specific knowledge that is to be learned by the students. To do this, the cognitive expert works with domain experts to extract and formalize the most important knowledge and techniques in a given domain. The experts are often provided by the client.

B. *Creating the scenario, the learner's profile and the toolbar*

Once the pieces of knowledge specific to the domain are clearly specified, the pedagogical expert assembles them in such a way as to identify the principal ones, define the pedagogical objectives of the SG and conceive a scenario that will allow learners to acquire them. This is where the SG research knot lies. Cognitive scientists have been working on didactic methods of teaching for years. They have managed to characterize different "learning moments" that make up the learning scenario [12] but it seems impossible to extract the essence of a "good" learning scenario. By collecting scenarios from SGs and courses that have proven to be effective, we hope to identify some successful "learning moment patterns" or at least identify best practice rules for building learning scenarios.

In addition to this issue, another challenge is raised by SGs: introducing fun into the learning scenario. Inspired by research on video gaming, film scripting and the art of storytelling, we have characterized a number of "entertaining moments" in the same way we identified "learning moments". The global SG scenario is then defined as the combination of an entertaining scenario and a learning scenario one on top of the other. As shown in fig. 2, the global SG scenario is then broken down into modules.

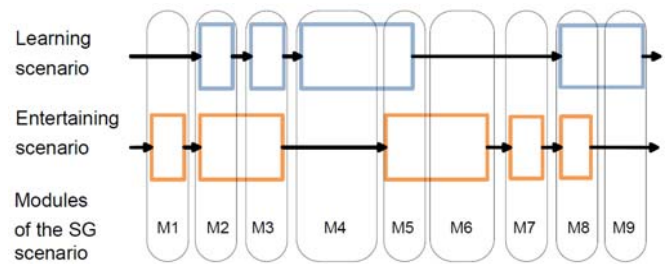

Fig. 2 Learning and entertaining scenario

The three different scenarios involved will be written with the help of a scenario editing tool. If the SG supports adaptation, the expert will also have to specify the different paths that can be taken and the determining criteria. The experience we have acquired in our research laboratory (a dozen SGs designed, tested and currently used in classrooms to train future engineers) leads us to identify several different types of scenarios (board, adventure, investigation game…). Our objective is to collect frequent SG types that could be used as a base to build on. For each module of the scenario, the expert then has to specify the sequence of activities the learner will have to follow.

Simple activity editors could be set up with editing interfaces, allowing pedagogical experts to create their own activities with a generic model. A number of micro-activity editors have already been put together for puzzles, crosswords, voting systems, brainstorming sequences, etc. These micro-activity editors are currently being collected and normalized so they can be integrated into any SG. Any activity, whether it be a micro-activity, an activity or a module (sequence of activities), is described by a metadata file that explains its use characteristics. Each activity can then be linked to several activity editors. The editors will be chosen in accordance with the pedagogical experts' experience with programming and the degree of adaptation they need. Let us illustrate this concept with a crossword puzzle: non-programmer pedagogical experts will choose a crossword editor with a simple editing interface that allows them to change the background color and the vocabulary used for the crossword whereas, if needed, computer scientists will choose a crossword editor that gives them access to the source code and thereby enables them to change all the elements of the cross word activity (presentation aspects, scoring techniques, words used, images…). The activity editors are therefore also tagged with metadata that specify the activities they can be applied to, the level of adaptation they offer and the minimum programming knowledge required for using them.

The different activity editors are placed in a library and accessed though a service oriented architecture (SOA) [13]. As Daniel Schneider concludes in his article [14], this is indeed the ideal method to access editors of different grain (applying to micro-activities, activities, modules…) that cannot be hierarchically classified.

The pedagogical expert then has to define the learner's profile by indicating the different pedagogical objectives to achieve and the mechanism for calculating the scores of each activity. He can also add extra indicators to measure the learner's behavior (paths followed, speed of action, number of times he clicked on a particular button…).

To finalize the conception, the expert has to edit the toolbar with the different tools accessible to the learner all through the game (help module, access to the history of game, pedagogic tools…).

### C. Editing the storyboard and specifying the virtual world and characters

The scenario is then structured in acts and scenes and enhanced with elements that make the SG fun (story, characters, intrigue, quest…). We propose to use theatre vocabulary to distinguish these activity sequences.

- Act: The SG is usually split into acts of approximately the same size (as concerns time and information). Each act is dedicated to one pedagogical objective.
- Scene: Subdivisions of an act in which there are no modification of characters or environment. A scene usually corresponds to a game interaction.

The scenario is therefore structured into logical sequences of activities that will be executed in order by the learner. Fig. 3 shows an example of an XML Act file. A raw XML description file of this type is not very easy to handle for a pedagogical expert. Our objective is therefore to provide an editing tool with a visual interface adapted to SGs. The interoperability will be insured by the use of existing specifications such as IMS-LD [15] and editors like Reload [16] or MOT+ [17].

```
<scenes>
    <scene num="1"    prec="0"    chemin="scenes/scene1.swf"    description="Roland Ketteur devant le BM"
            condition="&score>15" suiv1="2" suiv2="1"/>

        <!-- si la condition est satisfaite (la variable score est bien supérieure à 15)
        on passe à la scène suiv1 sinon on passe à suiv2 donc on reste dans la meme scene 1

    <scene num="2"    prec="1"    chemin="scenes/scene2.swf"    description="Roland Ketteur devant le BM"
            condition1="&score<10" suiv11="1" condition2="&score>15" suiv21="3" suiv22="2"/>

        <!-- si condition1 est satisfaite on passe à la scène suiv1
        sinon on teste condition2 si vrai on passe à suiv21 sinnon à suiv22
</scenes>
```

Fig. 3 Example of an XML storyboard file

Once the scenario has undergone all these modifications, it is called a storyboard. The artistic director then has to describe all the visual, sound and other presentation aspects. The ideal solution would be to use a communication charter to maximize the chances of having clear, precise specifications for the various specialists (graphic designers, actors, film directors, sound managers…)

As a result of this conception phase, the mock-up model of the SG, composed of the storyboard and the different specifications of the gaming aspects, contains all the necessary information to start production.

### D. Pedagogical quality control

To minimize the testing phases, we intend to set up a pre-evaluation of the SG before it is actually produced. A first set of tests can be run on the storyboard graph to make sure there are no dead-end paths and that all paths insure that the learners acquire the main pedagogical objectives. For a more thorough testing, we can model virtual players that will act in accordance to their level of knowledge and their specific behavioral profile (curious, prudent, hasty, confident…)[18][19]. For the time being, this method exists only for board game type SGs that have a very formalized and simple structure, but it should be applicable to other types. The objective of these simulations is to statistically evaluate the SG in terms of pedagogical gain. With this system, we hope to be more efficient then when testing is only done on real people at the end of the production, which usually results in going through the production chain again.

## V. CONCLUSION AND PERSPECTIVES

In this article, we propose a production chain model to efficiently conceive and produce SGs that are certified for their educational value and fun qualities. We propose various tools to help the human actors do their tasks in a rapid and efficient way. The idea of activity editors that will allow activities to be generically created is also brought up. For each step of the SG production chain, we have presented the different tools that presently exist. We now have to determine to what extent they may be used and modified to answer our particular needs. We will also have to determine a format for the different material and artifacts (documents, data base, XML…) used by these tools so that they can be integrated into the global interoperable structure.